\documentclass{elsarticle}
\usepackage{amssymb}

\journal{Neurocomputing}
\usepackage{graphicx}
\usepackage{epstopdf}
\usepackage{subfigure}
\usepackage{amsmath}
\usepackage{amssymb}
\usepackage{algorithm}
\usepackage{algpseudocode}
\usepackage{multirow}
\usepackage{multicol}
\usepackage[flushleft]{threeparttable}
\usepackage{balance}
\usepackage{adjustbox}
\usepackage[labelsep=period]{caption}

\makeatletter
\def\ps@pprintTitle{%
 \let\@oddhead\@empty
 \let\@evenhead\@empty
 \def\@oddfoot{}%
 \let\@evenfoot\@oddfoot}
\makeatother
\begin{document}
\setlength{\parindent}{2em}
\begin{frontmatter}

\title{\textbf{Zero-Shot Learning \\with Multi-Battery Factor Analysis}}

\author{Zhong Ji\textsuperscript{a}, Yuzhong Xie\textsuperscript{a}, Yanwei Pang\textsuperscript{a}, Lei Chen\textsuperscript{a}, Zhongfei Zhang\textsuperscript{b}}
\address{\textsuperscript{a}School of Electronic Information Engineering, Tianjin University, Tianjin, 300072, China}
\address{\textsuperscript{b}Department of Computer Science, State University of New York, Binghamton, NY 13902, USA}
\begin{abstract}

\par\setlength\parindent{1em}
Zero-shot learning (ZSL) extends the conventional image classification technique to a more challenging situation where the test image categories are not seen in the training samples. Most studies on ZSL utilize side information such as attributes or word vectors to bridge the relations between the seen classes and the unseen classes. However, existing approaches on ZSL typically exploit a shared space for each type of side information independently, which cannot make full use of the complementary knowledge of different types of side information. To this end, this paper presents an MBFA-ZSL approach to embed different types of side information as well as the visual feature into one shared space. Specifically, we first develop an algorithm named Multi-Battery Factor Analysis (MBFA) to build a unified semantic space, and then employ multiple types of side information in it to achieve the ZSL. The close-form solution makes MBFA-ZSL simple to implement and efficient to run on large datasets. Extensive experiments on the popular AwA, CUB, and SUN datasets show its significant superiority over the state-of-the-art approaches.
\end{abstract}


\end{frontmatter}

\section{Introduction and Related Work}
\par\setlength\parindent{1em}
Zero-shot learning (ZSL) aims at solving the problem when the new test image categories are not seen in the training samples \cite{Palatucci_nips_09}. Different from the open set recognition and novelty detection which only distinguish abnormalities in the testing data, ZSL seeks to classify the unseen testing classes \cite{Pimentel_sp_14}. This is a practical problem setting in image classification, as there are thousands of categories of objects we intend to recognize, but only a few of them may have been appropriately annotated. Consequently, it is more challenging than the conventional image classification problems. The key ideas of ZSL are to choose better side information (also known as modalities) and to develop an effective common semantic space. The side information provides a bridge to transfer knowledge from the seen classes for which we have training data to the unseen classes for which we do not, and the common space offers a fusion feasibility for the visual features and the side information.

\par\setlength\parindent{1em}
Two types of commonly used side information in ZSL are attributes \cite{Lampert_cvpr_09,Lampert_pami_14,Deng_eccv_14,Paredes_icml_15} and word vectors \cite{Pennington_emnlp_14}, \cite{Mikolov_nips_13}. Particularly, attributes act as intermediate representations shared across multiple classes, indicating the presence or absence of several predefined properties. Direct attribute prediction (DAP) \cite{Lampert_cvpr_09} is one of the first efforts to exploit the attributes to ZSL. It learns attribute-specific classifiers with the seen data and infers the unseen class with the learned estimators. However, attribute-based approaches suffer from a poor scalability as the attributes ontology for each class is generally manually defined. Word-vector-based approaches \cite{Norouzi_iclr_14,Socher_nips_13,Elhoseiny_iccv_13,Frome_nips_13} avoid this limitation since word vectors are extracted from a linguistic corpus with neural language models such as GolVe \cite{Pennington_emnlp_14} and Word2Vec \cite{Mikolov_nips_13}. Therefore, word vectors have become another popular side information in ZSL. For instance, Socher \textit{et al.} \cite{Socher_nips_13} construct a two layer neural network to project images into the word vector space. In \cite{Frome_nips_13}, Frome \textit{et al.} present a deep visual-semantic embedding model with a hinge loss function, which trains a linear mapping to link the image visual space to the word vector space.

\par
\begin{figure}

	\subfigure[Conventional approaches]{
		\begin{minipage}[b]{\textwidth}
		\centering
		\label{fig1_a}
		\includegraphics[width=\linewidth]{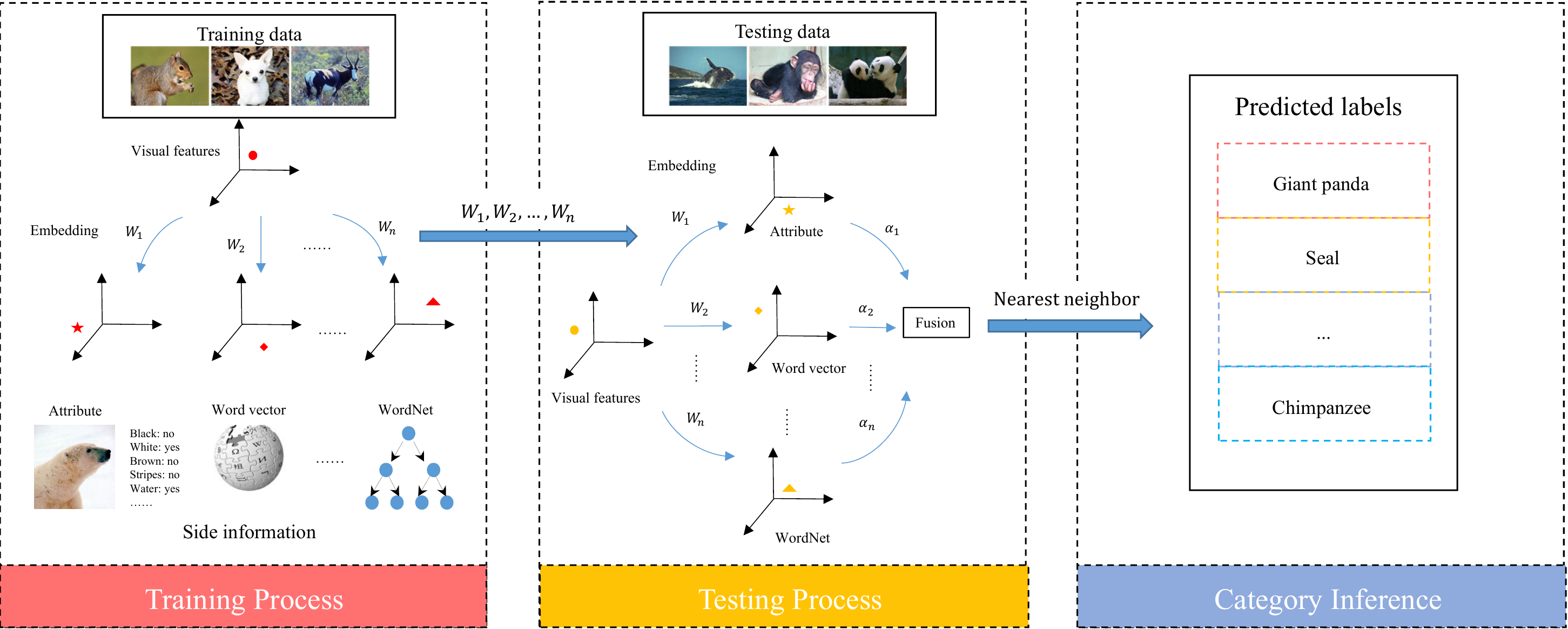}
		\end{minipage}
	}
	\subfigure[MBFA-ZSL approach]{
		\begin{minipage}[b]{\textwidth}
		\centering
		\label{fig1_b}
		\includegraphics[width=\linewidth]{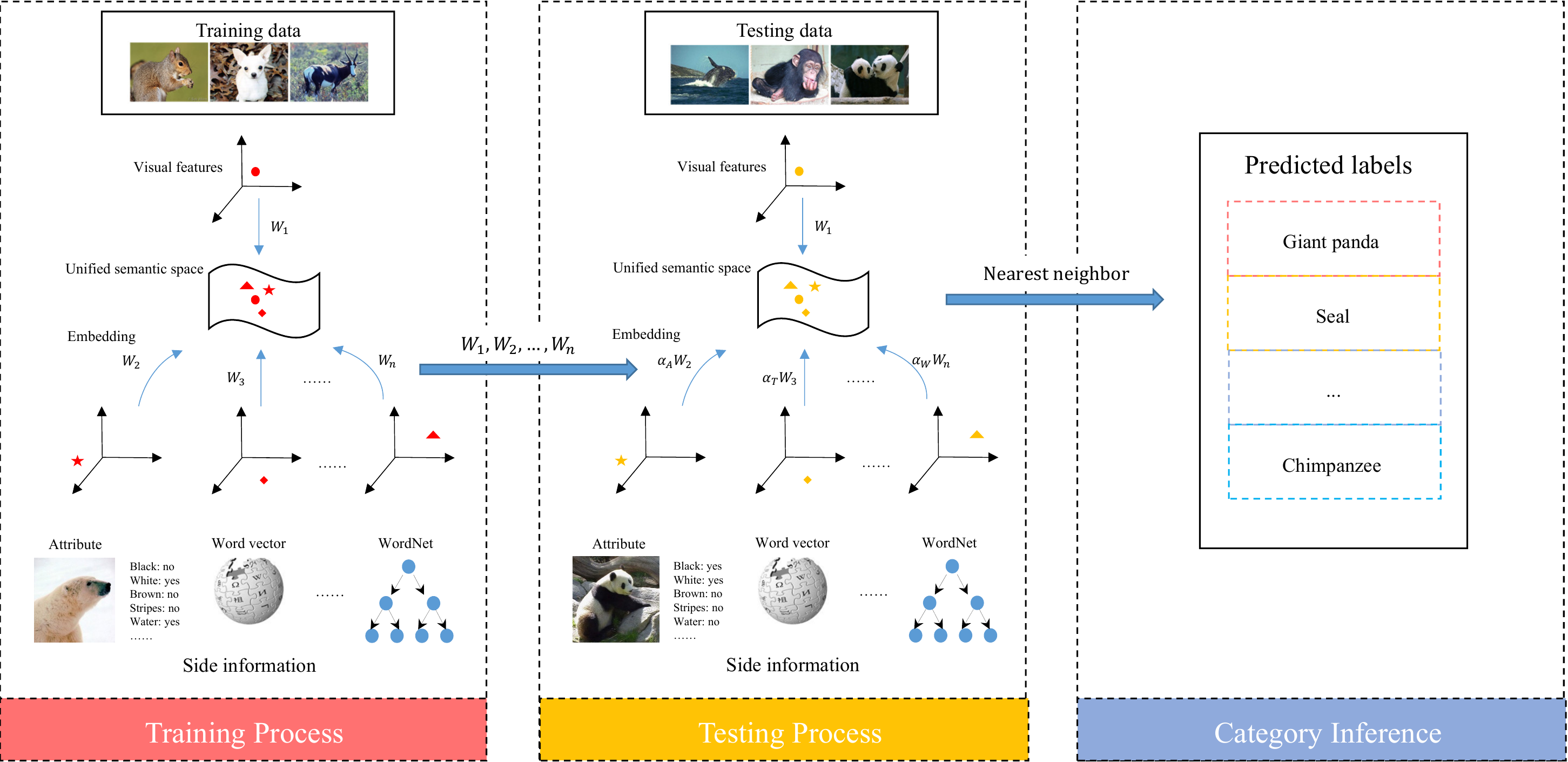}
		\end{minipage}
	}
	\caption{The comparative illustration of the proposed MBFA-ZSL and the conventional approaches. (a) Conventional approaches embed the visual features to each type of side information in its own semantic space independently, (b) MBFA-ZSL employs multiple types of side information in a unified space.}
	\label{fig1}
\end{figure}

\setlength\parindent{1em}
Besides attributes and word vectors, some other side information, such as WordNet \cite{Akata_cvpr_15}, visual prototypical concepts \cite{Jetley_bmvc_15}, class co-occurrence statistics \cite{Mensink_cvpr_14}, is also applied in ZSL. Further, since different types of side information captures different aspects of the structure of the semantic space, several studies have been made to combine them to achieve higher classification performance \cite{Akata_cvpr_15},\cite{Fu_pami_15},\cite{Fu_eccv_14}. For example, in \cite{Akata_cvpr_15}, Akata \textit{et al.} first learn the joint embedding weight matrices corresponding to different types of side information, then perform a grid search over the coefficients on a validation set to get the joint compatibility model. In \cite{Fu_pami_15}, semantic projections are trained for attributes and word vectors independently, followed by a transductive multi-view semantic embedding space to alleviate the projection domain shift problem. These efforts demonstrate that different types of side information complement each other and construct a better embedding space for knowledge transfer. However, although multiple types of side information are utilized, they still exploit each type of side information in its own semantic space independently, and then just combine the predicted scores together \cite{Akata_cvpr_15}, \cite{Fu_pami_15}. This cannot make full use of the complementary knowledge of different types of side information. A more efficient and robust solution is to investigate multiple types of side information in a unified space. Unfortunately, to the best of our knowledge, there has been little previous work exploiting this idea. To this end, we present a novel approach called MBFA-ZSL to employ multiple types of side information in a unified space, as shown in Fig. \ref{fig1}.

\par\setlength\parindent{1em}
It is worth highlighting several aspects of the proposed MBFA-ZSL approach. (1) It develops an advanced multi-view embedding algorithm named Multi-Battery Factor Analysis (MBFA), which extends Tucker's Inter-Battery Factor Analysis (IBFA) \cite{Tucker_IBFA_58}. (2) As far as we know, it represents one of the first attempts that embeds both the image visual features and multiple types of side information into one unified semantic space, which fully utilizes the interrelations among different types of information. (3) The close-form solution makes it simple to implement and efficient to run on large datasets. (4) Extensive experiments on popular datasets demonstrate its significant superiority over the existing state-of-the-art approaches.
\par\setlength\parindent{1em}
The reminder of this paper is structured as follows. Section 2 introduces the proposed Multi-Battery Factor Analysis (MBFA) algorithm, and Section 3 describes the proposed MBFA-ZSL approach in detail. Experimental results are presented in Section 4, and conclusions are drawn in the final section.

\section{Multi-Battery Factor Analysis}
\par\setlength\parindent{1em}
Multi-Battery Factor Analysis (MBFA) is developed to provide a unified semantic space for both the visual features and multiple types of side information. It originates from Tucker's Inter-Battery Factor Analysis (IBFA) \cite{Tucker_IBFA_58}, which transforms two modalities into a shared space where they are not only well explained but also as much correlated as possible with each other. Thus, we first briefly introduce the IBFA algorithm, and then extend it to a multi-view version, i.e., MBFA.

\par\setlength\parindent{1em}
Given a set of $N$ instances from two modalities, $\mathbf{X}_1=[\mathbf{x}_{11},...,\mathbf{x}_{1N}]\in{\mathbb{R}^{p_1\times{N}}}$ and $\mathbf{X}_2=[\mathbf{x}_{21},...,\mathbf{x}_{2N}]\in{\mathbb{R}^{p_2\times{N}}}$, where $p_1$ and $p_2$ are their dimensionalities, respectively. With $\mathbf{X}_1$, $\mathbf{X}_2$ centered, IBFA finds two projection matrices, $\mathbf{W}_1$ and $\mathbf{W}_2$, by the following constrained maximization:
\begin{equation} \label{eq:IBFA}
\begin{aligned}
& \max_{\mathbf{W}_1,\mathbf{W}_2}
& & \mathrm{tr}(\mathbf{W}_1^T\mathbf{X}_1\mathbf{X}_2^T\mathbf{W}_2), \\
& \mathrm{s.t.}
& & \mathbf{W}_1^T\mathbf{W}_1=\mathbf{I}, \; \mathbf{W}_2^T\mathbf{W}_2=\mathbf{I}.
\end{aligned}
\end{equation}
where $\mathbf{I}$ is an identity matrix. IBFA maximizes the total covariance between the two modalities, which can be seen plainly by rewriting $\mathrm{tr}(\mathbf{W}_1^T\mathbf{X}_1\mathbf{X}_2^T\mathbf{W}_2)$ as $\sum_{i=1}^{N}{\mathbf{W}_1^T\mathbf{x}_{1i}\mathbf{x}_{2i}^T\mathbf{W}_2}$. With the Lagrange multiplier method, (\ref{eq:IBFA}) can be solved analytically through the eigenvalue decomposition.
\par\setlength\parindent{1em}
Compared with Canonical Correlation Analysis (CCA) \cite{Hotelling_CCA_36}, (\ref{eq:IBFA}) can be rewritten as:
\begin{equation} \label{eq:IBFAv2}
\begin{aligned}
& \max_{\mathbf{W}_1,\mathbf{W}_2}
& & (\mathrm{corr}(\mathbf{W}_1^T\mathbf{X}_1, \mathbf{W}_2^T\mathbf{X}_2) \cdot \sqrt{\mathrm{var}(\mathbf{W}_1^T\mathbf{X}_1)} \cdot \sqrt{\mathrm{var}(\mathbf{W}_2^T\mathbf{X}_2)}), \\
& \text{s.t.}
& & \mathbf{W}_1^T\mathbf{W}_1=\mathbf{I}, \; \mathbf{W}_2^T\mathbf{W}_2=\mathbf{I}.
\end{aligned}
\end{equation}
where $\mathrm{corr}(\mathbf{a},\mathbf{b})$ denotes the Pearson correlation, and $\mathrm{var}(\mathbf{a})=\mathbf{a}^T\mathbf{a}$ is the variance. It can be seen from (\ref{eq:IBFAv2}) that IBFA attempts to capture both the correlation and variation of $\mathbf{X}_1$ and $\mathbf{X}_2$. This is different from CCA that only aims at maximizing their correlation. In particular, the maximized correlation and variance in (\ref{eq:IBFAv2}) depict the relationship between $\mathbf{X}_1$ and $\mathbf{X}_2$ and strengthen their own discriminant capabilities, respectively.
\par\setlength\parindent{1em}
To broaden IBFA to a multi-view scenario, we develop the MBFA algorithm on the basis of IBFA. Given a set of $N$ instances from $c$ modalities, $\mathbf{X}_i=[\mathbf{x}_{i1},...,\mathbf{x}_{iN}]\in{\mathbb{R}^{p_i\times{N}}}, i=1,\dots,c$, where $p_i$ denotes the dimensionality, with $\mathbf{X}_i$ centered, the objective function of MBFA is expressed as: 
\begin{equation} \label{eq:MBFA}
\begin{aligned}
& \max_{\mathbf{W}_1,\mathbf{W}_2,\dots,\mathbf{W}_c}
& & \sum_{\substack{i,j=1\\ i \neq j}}^{c}{\mathrm{tr}(\mathbf{W}_i^T\mathbf{X}_i\mathbf{X}_j^T\mathbf{W}_j)}. \\
& \mathrm{s.t.}
& & \mathbf{W}_i^T\mathbf{W}_i=\mathbf{I}, \; i=1,\dots,c, c \geq 2.
\end{aligned}
\end{equation}
Similar to IBFA, MBFA tries to find a set of projection matrices that maximize the total covariance in the common space. Equation (\ref{eq:MBFA}) can be rewritten as:
\begin{equation} \label{eq:MBFAv2}
\begin{aligned}
& \max_{\mathbf{W}}
& & \mathrm{tr}(\mathbf{W}^T\mathbf{M}\mathbf{W}), \\
& \mathrm{s.t.}
& & \mathbf{W}^T\mathbf{W}=\mathbf{I}.
\end{aligned}
\end{equation}
where $\mathbf{W}$ and $\mathbf{M}$ are as follows:
\begin{equation} \label{eq:W}
\mathbf{W} = [\mathbf{W}_1^T, \mathbf{W}_2^T, \dots, \mathbf{W}_c^T]^T.
\end{equation}
\begin{equation}
\mathbf{M} =
\left[ \begin{array}{cccc}
\mathbf{M}_{11} & \mathbf{M}_{12} & \ldots & \mathbf{M}_{1c} \\
\mathbf{M}_{21} & \mathbf{M}_{22} & \ldots & \mathbf{M}_{2c} \\
\vdots & \vdots & \ddots & \vdots \\
\mathbf{M}_{c1} & \mathbf{M}_{c2} & \ldots & \mathbf{M}_{cc} \\
\end{array} \right],
\mathbf{M}_{ij} = \left\{ \begin{array}{ll}
	0, & i=j \\
	\mathbf{X}_i\mathbf{X}_j^T, & i \neq j.
\end{array} \right.
\end{equation}
Equation (\ref{eq:MBFAv2}) can be solved via the eigenvalue decomposition; thus each projection matrix $\mathbf{W}_i$ can be obtained. It is obvious that IBFA can be considered as a special case of MBFA when $c$ is 2. 

\par\setlength\parindent{1em}
In addition, the main difference between MBFA and Multi-view Canonical Correlation Analysis (MCCA) \cite{Gong_ijcv_14}, \cite{Rupnik_sikdd_10} is worth highlighting. Both of them find a set of linear transformations to project multiple modalities into one common space. However, MCCA seeks to maximize the total correlation in the common space, whereas MBFA maximizes the total covariance, which is equivalent to maximize the total correlation and variance simultaneously. To the best of our knowledge, there is no previous work using MCCA on ZSL. In this paper, we also implement MCCA on ZSL as a comparative approach (we call this approach as MCCA-ZSL).

\section{Zero-Shot Learning with MBFA}
\par\setlength\parindent{1em}
In a ZSL setting, we are given $N_s$ labeled training instances $\mathcal{S}=\{\mathbf{X}, \mathbf{Y}^k, \mathbf{z}\}$ and $N_u$ unlabeled testing instances $\mathcal{U}=\{\mathbf{\widetilde{X}}, \mathbf{\widetilde{Y}}^k, \mathbf{\widetilde{z}}\}$. $\mathbf{X} \in \mathbb{R}^{p\times{N_s}}$ and $\mathbf{\widetilde{X}} \in \mathbb{R}^{p\times{N_u}}$ are the $p$-dimensional visual feature vectors of training and testing instances respectively. $\mathbf{z}$ and $\mathbf{\widetilde{z}}$ are the seen and unseen class label vectors, and $\mathbf{z} \cap \mathbf{\widetilde{z}} = \varnothing$. We have $K$ different types of side information, $\mathbf{Y}^k \in \mathbb{R}^{q_k\times{N_s}}$ and $\mathbf{\widetilde{Y}}^k \in \mathbb{R}^{q_k\times{N_u}}$ denote the $k$-th type of $q_k$-dimensional side information for training and testing datasets respectively. Note that for the testing dataset, $\mathbf{\widetilde{Y}}^k$ is missing as testing instances are unlabeled. The task of ZSL is to predict the class labels $\mathbf{\widetilde{z}}$.

\par\setlength\parindent{1em}
The proposed MBFA-ZSL algorithm mainly contains the following two steps: 

\par\setlength\parindent{1em}
\emph{Step 1: Build a MBFA space with the seen data.} The MBFA algorithm provides an unified semantic space $\mathbb{Z}$ for different types of side information as well as the visual features. With the seen images together with the side information, we can train the MBFA model to obtain a set of projection matrices $\mathbf{W}_i$ $(i=1, \dots, c)$, where $c$ is the sum of all the types of side information and the visual features, such that $c=K+1$. For example, if we use both attributes and word vectors as the side information, then $c$ is 3.

\par\setlength\parindent{1em}
\emph{Step 2: Unseen category Inference.} With the projection matrix $\mathbf{W}_1$ learned from the seen data, the unseen image features $\mathbf{\widetilde{x}}_j \in \mathbf{\widetilde{X}}$ can be embedded into the common space $\mathbb{Z}$ by $\theta(\mathbf{\widetilde{x}}_j)=\mathbf{W}_1^T\mathbf{\widetilde{x}}_j$. Typically, the unseen category of $\mathbf{\widetilde{x}}_j$ can be inferred by searching for the nearest output embedding vector that corresponds to one of the unseen classes, if there is only single side information available in $\mathbb{Z}$. Since there are multiple types of side information used in the MBFA-ZSL approach, we introduce a multi-modality fusion method to predict the unseen category of the $\mathbf{\widetilde{x}}_j$ with:
\begin{equation} \label{eq:inference}
\begin{aligned}
l^*=\underset{l}{\operatorname{argmax}}\bigg[\sum_{k=1}^K{\alpha_{k}\mathrm{sim}\Big(\theta(\mathbf{\widetilde{x}}_j), \varphi_k(\mathbf{\widetilde{y}}_l^k)\Big)}\bigg],l=1,2,\dots,n,
\end{aligned}
\end{equation}
where $\alpha_{k}$ is a weight associated with each type of side information, which can be determined by a grid search on the validation set. Each type of side information that corresponds to the $l$-th unseen class is denoted as $\mathbf{\widetilde{y}}_l^k$, and it can be embedded into the common space $\mathbb{Z}$ by $\varphi_k(\mathbf{\widetilde{y}}_l^k)=\mathbf{W}_{k+1}^T\mathbf{\widetilde{y}}_l^k$. The similarity between two vectors can be represented as the common distance measurements, such as dot product similarity and Euclidean distance. In this paper, the cosine distance is utilized, i.e., $\mathrm{sim}(\mathbf{a}, \mathbf{b})=\mathbf{a}^T\mathbf{b}/(\|\mathbf{a}\| \cdot \|\mathbf{b}\|)$

\par\setlength\parindent{1em}
Moreover, MBFA-ZSL has an explicit, close-form solution, which makes it simple to implement and efficient to run on large datasets. \textbf{Algorithm 1} outlines the procedures of the proposed MBFA-ZSL approach.

\par\setlength\parindent{1em}
\algnewcommand\algorithmicinput{\textbf{Input:}}
\algnewcommand\INPUT{\item[\algorithmicinput]}
\algnewcommand\algorithmicoutput{\textbf{Output:}}
\algnewcommand\OUTPUT{\item[\algorithmicoutput]}

\begin{algorithm}
\caption{MBFA-ZSL approach} \label{MBFA-ZSL}
\begin{algorithmic}[1]
\INPUT{ A labeled seen data set $\mathcal{S}=\{\mathbf{X}, \mathbf{Y}^k, \mathbf{z}\}$, an unlabeled unseen data set $\mathcal{U}=\{\mathbf{\widetilde{X}}, \mathbf{\widetilde{Y}}^k, \mathbf{\widetilde{z}}\}$, and the dimensionality $d$ of the unified embedding space.}
\OUTPUT{ Labels of the unseen data $\mathcal{U}$}.
\State Construct the covariance matrix $\mathbf{M}$ with the labeled visual features $\mathbf{X}$ and the corresponding side information $\mathbf{Y}^k$.
\State Solve the eigenvalue decomposition problem in (\ref{eq:MBFAv2}), and the eigenvectors corresponding to the largest $d$ eigenvalues form the projection matrices $\mathbf{W}$.
\State Learn the weight parameters of the category inference function (\ref{eq:inference}) in the validation set.
\State Project the unseen visual feature $\mathbf{\widetilde{X}}$ and the side information of the unseen classes into the unified space with projection matrices $\mathbf{W}$. 
\State Predict the labels of $\mathcal{U}$ with (\ref{eq:inference}).
\end{algorithmic}
\end{algorithm}

\section{Experimental Results and Discussion}
\subsection{Datasets and Settings}
\par
\begin{figure}
	\centering
	\subfigure[Example images of the AwA dataset]{
		\label{fig2_a}
		\includegraphics[width=\linewidth]{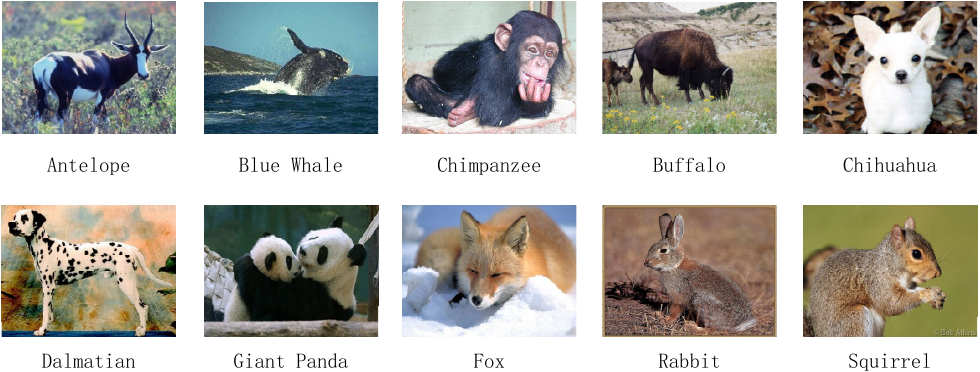}}
	\subfigure[Example images of the CUB dataset]{
		\label{fig2_b}
		\includegraphics[width=\linewidth]{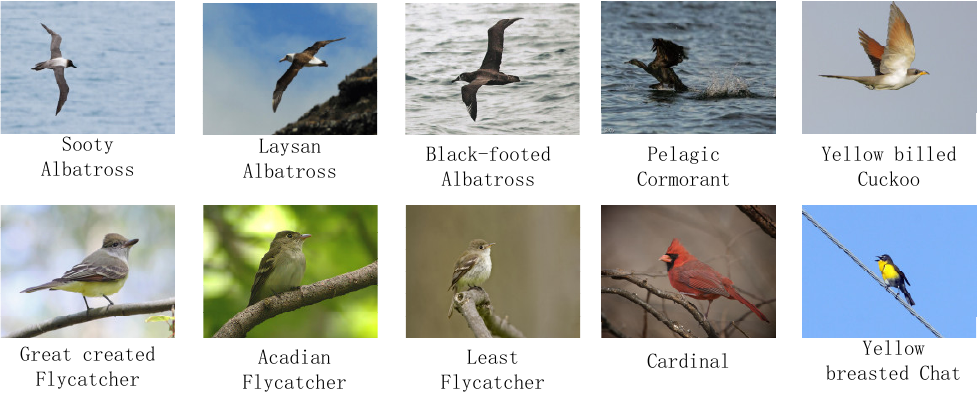}}
	\subfigure[Example images of the SUN dataset]{
		\label{fig2_c}
		\includegraphics[width=\linewidth]{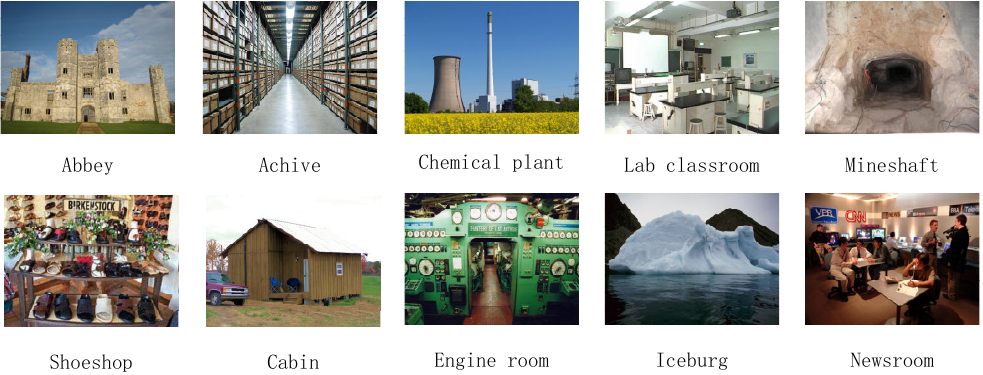}
	}
	\caption{Example images of the AwA, CUB, and SUN datasets.}
	\label{fig2}
\end{figure}

\setlength\parindent{1em}
We evaluate the proposed MBFA-ZSL approach on three publicly popular datasets: Animals with Attributes (AwA) \cite{Lampert_cvpr_09}, Caltech-UCSD-Birds-200-2011 (CUB) \cite{Wah_CUB}, and SUN Attribute \cite{Patterson_SUN}. Specifically, AwA is a collection of 30,475 images on 50 classes of animals, with 85 associated class-level attributes. We use the standard training/test (seen/unseen) split as that in \cite{Lampert_cvpr_09}, which chooses 40 classes for training and validation and 10 classes for testing. CUB provides 200 classes of birds (11,788 images), and each class is annotated with 312 attributes. Particularly, CUB is a much more challenging dataset in that it is designed for fine-grained image classification and contains more classes but fewer images. Similar to \cite{Akata_cvpr_15}, we use 150 classes as training and validation set, leaving 50 disjoint classes as test set. SUN Attribute dataset consists of 14,340 images from 717 scene categories, and each category is annotated with a taxonomy of 102 discriminate attributes. We adopt the popular training/test (seen/unseen) split as that in \cite{Jayaraman_nips_14}, which selects 707 classes for training and validation, and takes the remaining 10 classes as testing set. We cross-validate the parameters $\alpha_{k}$ in (\ref{eq:inference}). The example images in these datasets are shown in Fig. \ref{fig2}.

\par\setlength\parindent{1em}
On the AwA dataset, we use the VGG (very deep 19-layer CNN) features provided in \cite{AwA_web} as visual features. On the CUB and SUN dataset, we use a pre-trained VGG model to extract visual features \cite{Simonyan_iclr_15}. For each image, the 4,096 dimensional top-layer hidden unit activations (fc7) of VGG are taken as visual features.

\par\setlength\parindent{1em}
We use both the word vectors (T) and attributes (A) as the side information in MBFA-ZSL. Specifically, we train the Word2Vec model \cite{Mikolov_nips_13} on a corpus of Wikipedia documents to form 1000-D word vectors for the three datasets. Meanwhile, we use the attribute information provided by the datasets. The average per-class top-1 accuracy on the test sets is reported.

\subsection{Results on the AwA, CUB, and SUN datasets}
\par\setlength\parindent{1em}
We compare the proposed MBFA-ZSL with 7 state-of-the-art approaches as well as MCCA-ZSL, which utilize a range of side information. Among them, DAP \cite{Lampert_cvpr_09}, \cite{Lampert_pami_14},  ESZSL \cite{Paredes_icml_15}, and SSE-ReLU \cite{Zhang_iccv_15} only use attributes; LatEm \cite{Xian_cvpr_16} can make use of either word vectors or attributes; SJE \cite{Akata_cvpr_15}, AMP \cite{Fu_cvpr_15}, TMV-HLP \cite{Fu_pami_15} and MCCA-ZSL employ more than one type of side information. Different CNN visual features are applied in these approaches, such as GoogLeNet \cite{GoogLeNet}, Overfeat \cite{Overfeat}, and VGGNet-19 \cite{Simonyan_iclr_15}. Additionally, we also implement MBFA-ZSL and MCCA-ZSL in the situation where only attributes (A) or word vectors (T) are available.

\par\setlength\parindent{1em}
The performance of MBFA-ZSL are taken via ten times of cross validation. It should be noticed that when only T or A is avaliable, the single parameter $\alpha_{k}$ in (\ref{eq:inference}) does not need to be tuned, thus there is no standard deviation for the corresponding results. Furthermore, the standard deviations of some comparative results are absent as they are not avaliable in the original papers. The comparative results are summarized in Table \ref{performance}, from which we can observe that MBFA-ZSL achieves the amazingly best performance in all cases for all the three datasets. Besides, we also have the following observations: 

\begin{table*}[!h]
\centering
\caption{Performance Comparison (\%, mean$\pm$standard deviation) on the AwA, CUB, and SUN Datasets} \label{performance}
\begin{adjustbox}{max width=\textwidth}
\begin{tabular}{c|c|ccccccccc}
\hline
\multirow{2}{*}{Image feature} & \multirow{2}{*}{Approach} & \multicolumn{3}{c}{AwA} & \multicolumn{3}{c}{CUB} & \multicolumn{3}{c}{SUN} \\ \cline{3-11}
& & T & A & T+A & T & A & T+A & T & A & T+A \\
\hline
\multirow{2}{*}{GoogLeNet-22} & SJE \cite{Akata_cvpr_15} & 51.2 & 66.7 & 73.5* & 28.4 & 50.1 & 51.0* & - & - & - \\
& LatEm \cite{Xian_cvpr_16} & 61.1 & 71.9 & - & 31.8 & 45.5 & - & - & - & - \\ \hline
\multirow{2}{*}{Overfeat-8} & AMP \cite{Fu_cvpr_15} & - & - & 66.0 & - & - & - & - & - & - \\
& TMV-HLP \cite{Fu_pami_15} & - & - & 73.5 & - & - & 47.9 & - & - & - \\ \hline
\multirow{5}{*}{VGGNet-19} & DAP \cite{Lampert_pami_14} & - & 60.8 & - & - & - & - & - & 72.0 & - \\
& SSE-ReLU \cite{Zhang_iccv_15} & - & 76.3$\pm$0.8 & - & - & 30.4$\pm$0.2 & - & - & 82.5$\pm$1.3 & - \\
& ESZSL** \cite{Paredes_icml_15} & - & 74.6$\pm$3.7 & - & - & 50.8$\pm$0.4 & - & - & 84.5$\pm$1.4 & -\\
& MCCA-ZSL** & 65.8$\pm$1.7 & 74.9$\pm$0.3 & 75.3$\pm$1.8 & 32.1$\pm$0.3 & 45.8$\pm$0.2 & 46.4$\pm$0.7 & 59.5$\pm$1.7 & 82.8$\pm$0.5 & 85.1$\pm$1.5 \\
& MBFA-ZSL** & \textbf{72.5} & \textbf{77.8} & \textbf{79.9$\pm$0.7} & \textbf{32.4} & \textbf{51.7} & \textbf{52.2$\pm$0.4} & \textbf{61.5} & \textbf{85.0} & \textbf{87.4$\pm$0.2} \\
\hline 
\multicolumn{8}{l}{T, A represent attributes and word vectors, respectively.}\\
\multicolumn{8}{l}{*: additional WordNet hierarchies are used; **: our implementation.}\\
\end{tabular} 
\end{adjustbox}
\end{table*}

\par\setlength\parindent{1em}
(1) For AwA dataset, the second-best approaches are MCCA-ZSL, SSE-ReLU, and MCCA-ZSL in the cases of T, A, and T+A, respectively. MBFA-ZSL outperforms them in 6.7\%, 1.5\%, and 4.6\%, respectively. For CUB dataset, MBFA-ZSL outperforms the second-best approaches, MCCA-ZSL in 0.3\%, ESZSL in 0.9\%, and SJE in 1.2\% in the three cases, respectively. For SUN dataset, in the three cases, MBFA-ZSL outperforms the second-best approaches, MCCA-ZSL in 2.0\%, ESZSL in 0.5\%, and MCCA-ZSL in 2.3\%, respectively. These are very promising results. 

\par\setlength\parindent{1em}
(2) For MBFA-ZSL, the performance on AwA in T+A is better than those in T and A in 7.4\% and 2.1\%, respectively. On CUB, the performance in T+A is better than those in T and A in 19.8\% and 0.5\%, respectively. On SUN, the promotions are 25.9\% and 2.4\%, respectively. Similar observation can also be found in MCCA-ZSL. The excellent performance in T+A of MBFA-ZSL and MCCA-ZSL demonstrates that it is effective to embed multiple types of side information into a unified space. It also confirms that different types of side information complement each other in transferring knowledge. 

\par\setlength\parindent{1em}
(3) In the situation of ``T+A'', it can be found that MBFA-ZSL outperforms the others significantly. Take the AwA for example, the performance improvements of MBFA-ZSL over SJE, AMP and TMV-HLP are 6.4\%, 13.9\%, and 6.4\%, respectively. This demonstrates that embedding the visual features and multiple types of side information in a unified space is more promising than the conventional methods that projecting the visual features to each type of side information space independently at first, and then combining them together.

\par\setlength\parindent{1em}
(4) When only single type of side information is available, attributes often help achieve a higher accuracy than word vectors. This is due to that attributes are manually defined for a specific dataset, so they are able to describe category relationship of the dataset more effectively; nevertheless, word vectors are extracted from the corpus in an unsupervised manner, whose capacity is constrained by the size or specific domain of the corpus, as well as the polysemy issue. 

\par\setlength\parindent{1em}
(5) Interestingly, the performance on CUB is inferior to that on AwA and SUN. The reason may lie in the fine-grained characteristic of CUB. Both the visual appearance and the class names in it are similar to each other, which make it hard to recognize.

\par
\begin{figure}[!h]
	\centering
  	\includegraphics[width=3in]{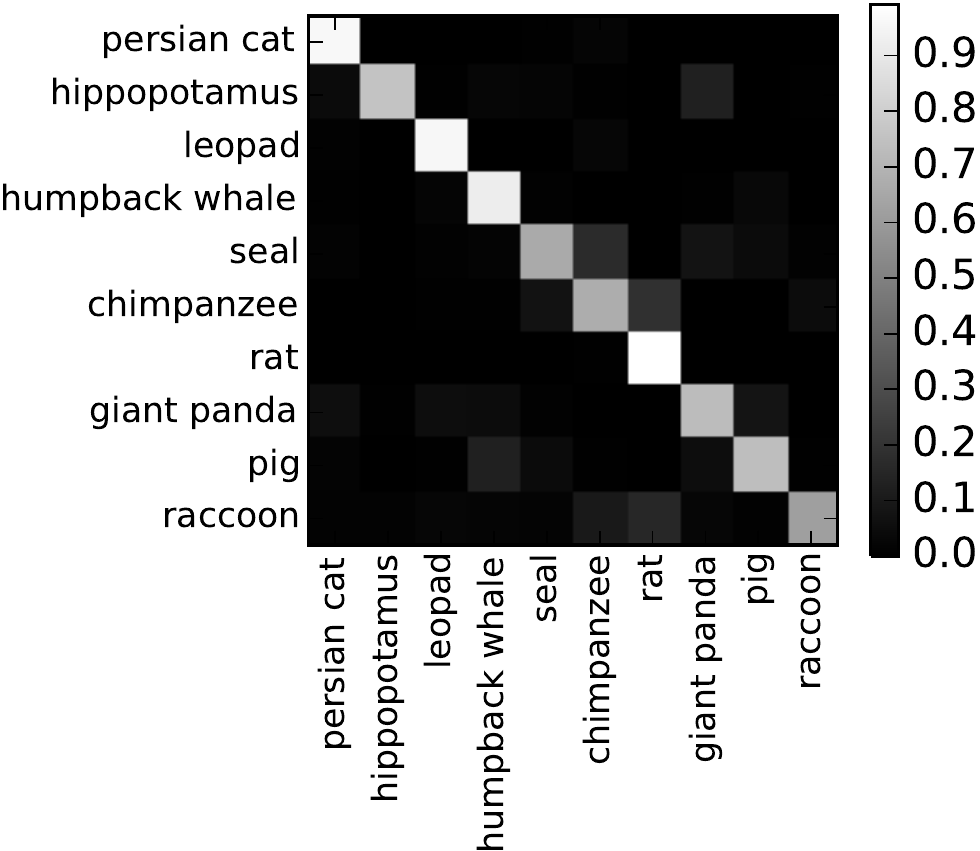}
  	\captionof{figure}{The confusion matrix between test classes of the AwA dataset.}
  	\label{confusionMat}
\end{figure}

\setlength\parindent{1em}
To clearly evaluate the performance of MBFA-ZSL on each class, we present the confusion matrix of AwA with T+A, as illustrated in Fig. \ref{confusionMat}. The diagonal elements denotes the correct prediction accuracy of each class, from which we can see that the proposed MBFA-ZSL can achieve relatively high performance on every class.

\begin{figure}
	\centering
	\subfigure[The AwA dataset]{
		\label{fig4_a}
		\includegraphics[width=0.5\linewidth]{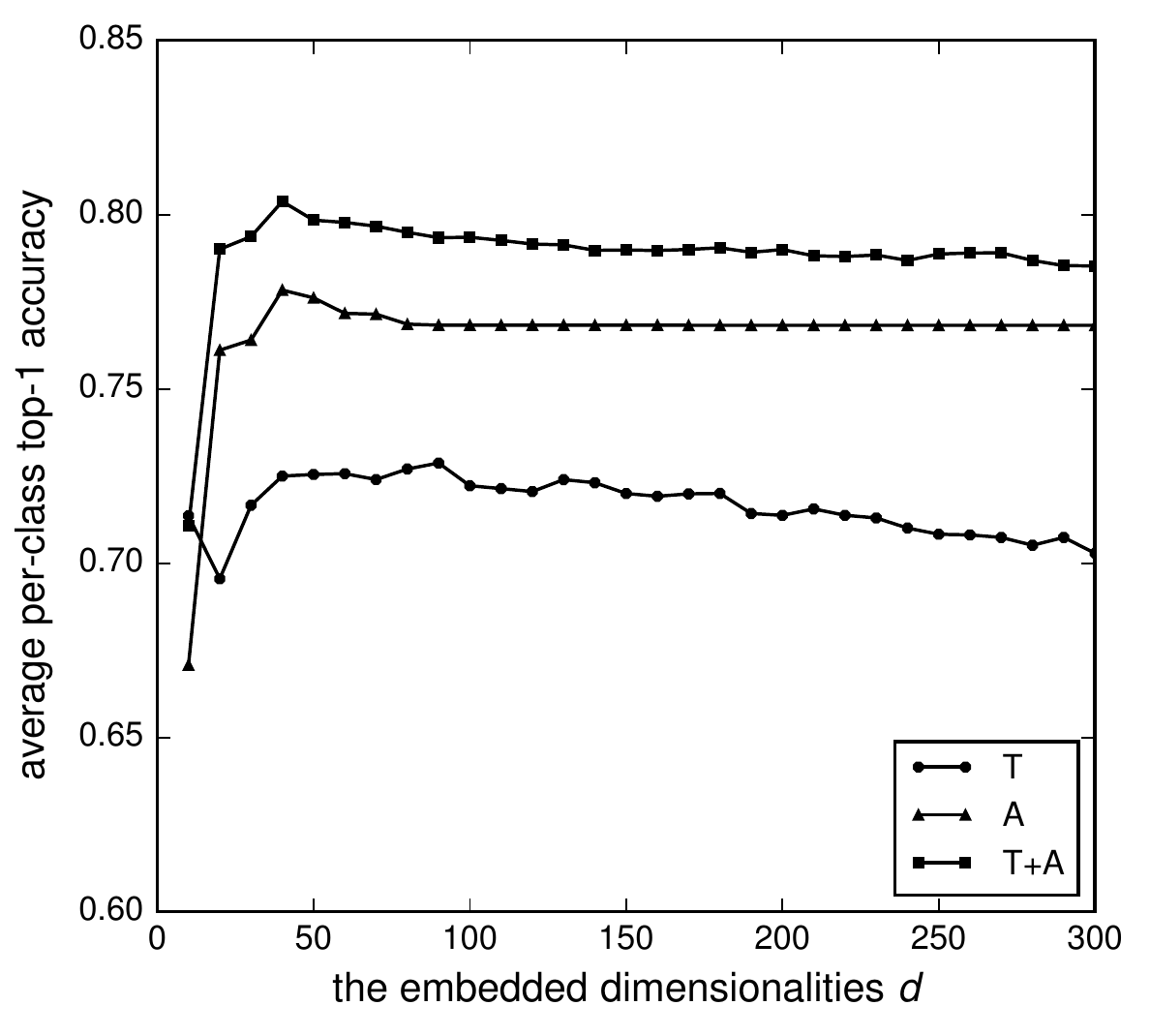}}
	\subfigure[The CUB dataset]{
		\label{fig4_b}
		\includegraphics[width=0.5\linewidth]{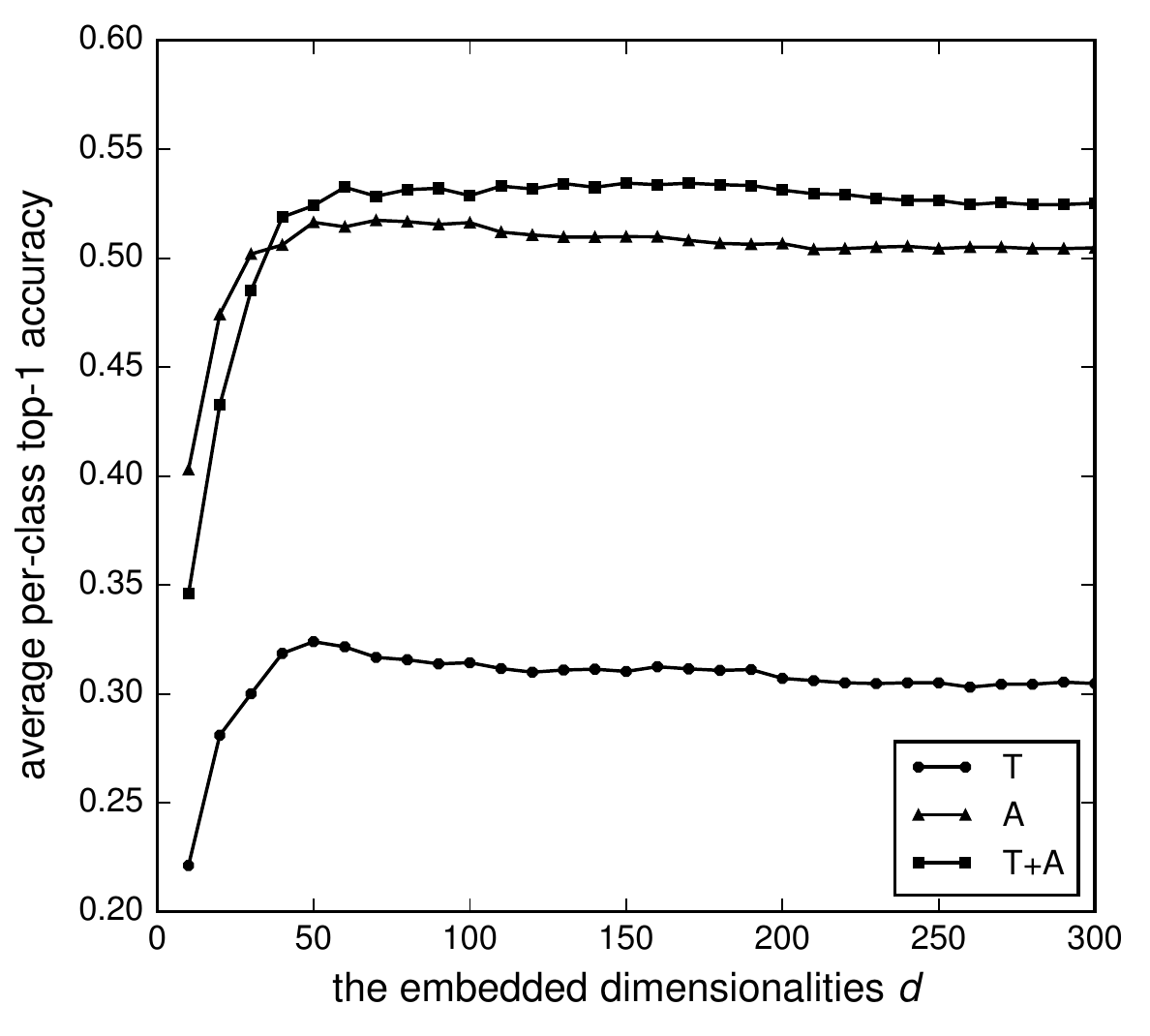}}
	\subfigure[The SUN dataset]{
		\label{fig4_c}
		\includegraphics[width=0.5\linewidth]{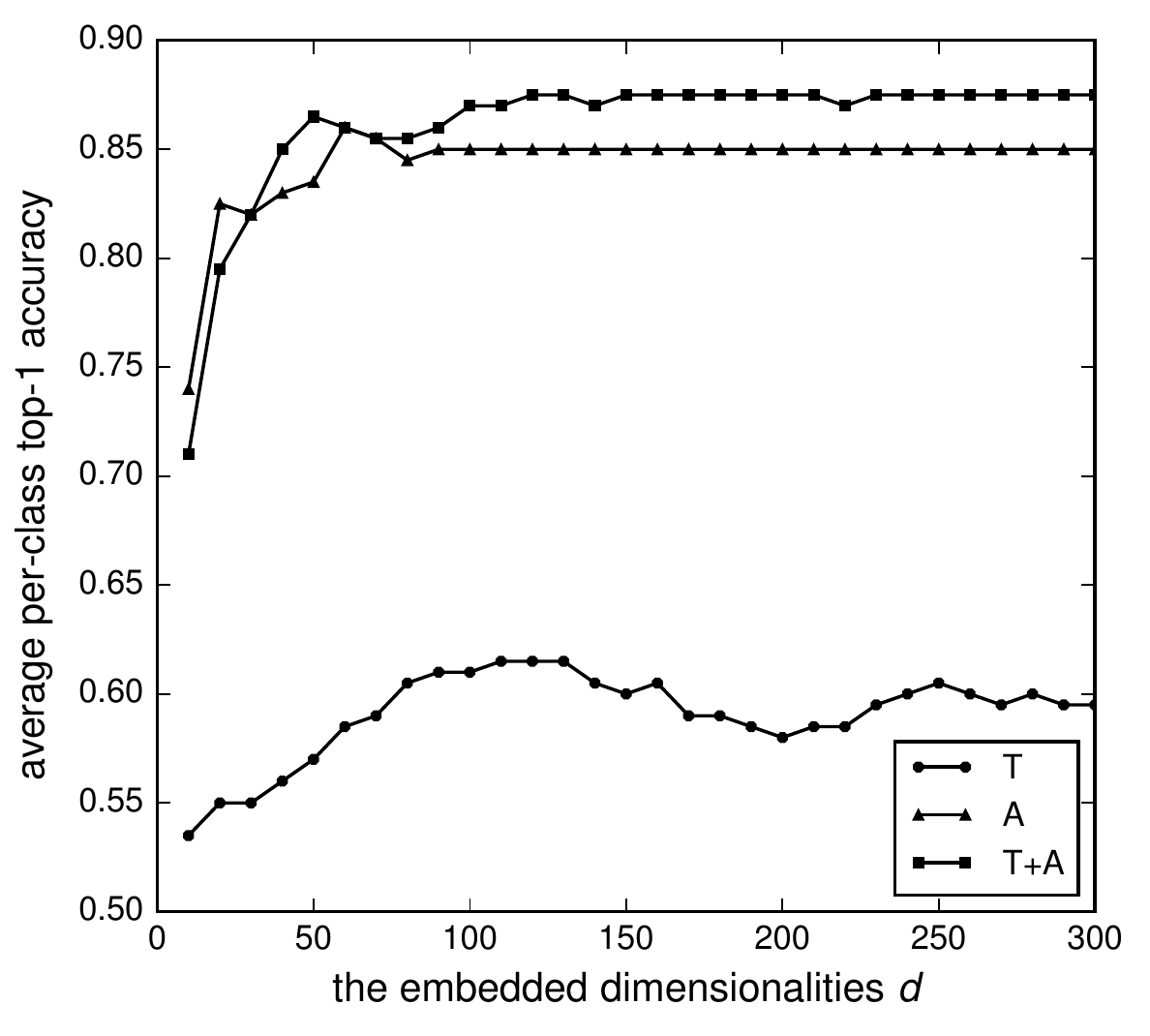}
	}
	\caption{The average per-class top-1 accuracy of MBFA-ZSL on AwA, CUB, and SUN datasets with respect to different settings and the embedded dimensionalities $d$. The types of side information are word vectors (T), attributes (A), and both of them (T+A).}
	\label{parameter}
\end{figure}

\subsection{Parameter Sensitivity}
\par\setlength\parindent{1em}
There are two types of parameters in MBFA-ZSL: the weights $\alpha_{k}$ in (\ref{eq:inference}) and the dimensionality $d$ of the unified space that multiple modalities are projected into. The weights are decided by the cross validation. The impact of dimensionalities $d$ is shown in Fig. \ref{parameter}. The optimal dimensionalities for AwA, CUB, and SUN are 40, 50, and 120, respectively. It can be observed that a higher dimensionality has no performance improvement.

\subsection{Speed Evaluation}
\par\setlength\parindent{1em}
Finally, as shown in Table \ref{speed}, we report the running times of the training and testing stages for AwA, CUB, and SUN, respectively. Our implementation is based on an unoptimized Matlab code. On our computer with i5 4590 CPU and 12G memory, the training times for the three datasets are 18.8s, 22.2s, and 17.4s, respectively. The test times on each image for the three datasets are 0.006ms, 0.010ms, and 0.019ms, respectively. Therefore, MBFA-ZSL is extremely efficient.

\begin{table*}[!h]
\centering
\caption{Training and Testing Times on the AwA, CUB, and SUN Datasets} \label{speed}
\begin{adjustbox}{max width=\textwidth}
\begin{tabular}{|c|c|c|c|}
\hline
{} & AwA & CUB & SUN \\ \hline
{Average training time for all the training data (s)} & 18.8 & 22.2 & 17.4 \\ \hline
{Average training time on each image (ms)} &0.8 & 2.5 & 1.2 \\ \hline
{Average testing time for all the testing data (ms)} & 39.0 & 29.5 & 3.8 \\ \hline
{Average testing time on each image (ms)} & 0.006 & 0.010 & 0.019 \\ \hline
\end{tabular} 
\end{adjustbox}
\end{table*}

\section{Conclusions}
\par\setlength\parindent{1em}
In this paper, we have proposed the MBFA-ZSL approach to projecting both the visual features and multiple types of side information into one unified semantic space to perform ZSL. It can also be applied to the situation where only a single type of side information is available. The results on the three popular datasets show its superior performance over the state-of-the-art approaches on the cases of utilizing A (attributes), T (word vectors), and A+T (attributes + word vectors) as an effective and efficient method. Moreover, it has a close-form solution.\\
\\
\textbf{Acknowledgements}
\par\setlength\parindent{1em}
This work was supported by the National Basic Research Program of China (973 Program) under Grant 2014CB340400, the National Natural Science Foundation of China under Grant 61271325, Grant 61472273, the Elite scholar Program of Tianjin University under Grant 2015XRG-0014, and the Research Program of Application Foundation and Advanced Technology of Tianjin under Grant 15JCYBJC17100.

\newpage

\end{document}